\documentclass[10pt, a4paper]{article}

\usepackage[final]{lrec2026} 
\usepackage{graphicx}
\usepackage{multirow}
\usepackage{booktabs}
\usepackage{float}
\usepackage{enumitem}
\usepackage{comment}

\title{Variation is the Norm: Embracing Sociolinguistics in NLP}

\name{Anne-Marie Lutgen$^{1}$, Alistair Plum$^{1}$, Verena Blaschke$^{2,3}$, \\
      {\bfseries \large Barbara Plank$^{2,3}$, Christoph Purschke$^{1}$}}

\address{$^{1}$University of Luxembourg, Esch-sur-Alzette, Luxembourg \\
         $^{2}$MaiNLP, LMU Munich, Germany \\
         $^{3}$Munich Center for Machine Learning (MCML), Munich, Germany \\
         \texttt{anne-marie.lutgen@uni.lu}}

\abstract{
In Natural Language Processing (NLP), variation is typically seen as noise and ``normalised away'' before processing, even though it is an integral part of language. Conversely, studying language variation in social contexts is central to sociolinguistics. We present a framework to combine the sociolinguistic dimension of language with the technical dimension of NLP. We argue that by embracing sociolinguistics, variation can actively be included in a research setup, in turn informing the NLP side. To illustrate this, we provide a case study on Luxembourgish, an evolving language featuring a large amount of orthographic variation, demonstrating how NLP performance is impacted. The results show large discrepancies in the performance of models tested and fine-tuned on data with a large amount of orthographic variation in comparison to data closer to the (orthographic) standard. Furthermore, we provide a possible solution to improve the performance by including variation in the fine-tuning process. This case study highlights the importance of including variation in the research setup, as models are currently not robust to occurring variation. Our framework facilitates the inclusion of variation in the thought-process while also being grounded in the theoretical framework of sociolinguistics. 
 \\ \newline  }

\begin{document}

\maketitleabstract

\section{Introduction}
In structural linguistics, variation is often perceived as a disturbing factor and discarded in grammatical descriptions to emphasise the constant and invariable elements of a language \cite{Berruto+2004+293+322}. With the advent of sociolinguistics, the study of language variation in social contexts \cite{wodak2011sociolinguistics} has become a central focus of linguistic research, mirroring the fundamental role of variation in language and the production of social meaning \cite{Eckert_2016}. 

Similarly to structural linguistics, variation in Natural Language Processing (NLP) is also often seen as noise (in the signal) and a practical nuisance (for processing) \cite{nguyen-etal-2021-learning}. We argue that sociolinguistic insight should be a part of the NLP research setup, since language variation and its role in constructing social meaning are not an exception but a characteristic of language and could therefore improve the performance of language models as well as their representation of linguistic diversity.

We develop a framework to combine the sociolinguistic classification of language variation with domains of application in NLP. This framework includes guidelines to understand the status and function of a linguistic entity (language or variety; see Section \ref{framework}) and the dimensions of variation linked to it. The technical dimension illustrates five steps in language modelling where variation has an impact. By combining these two sides, we achieve a precise understanding of the linguistic entity researched and how it is linked to the technical implementation of modelling. This also leads to practical solutions for how to handle variation in a processing pipeline, as a (socio)linguistic description of the researched entity will identify where problems on the technical side may occur.

We illustrate the use and effectiveness of our framework on a case study of orthographic variation in Luxembourgish and how it impacts the performance of fine-tuned classification models. We conduct an experiment where we fine-tune a Luxembourgish BERT model \cite{lothritz-etal-2022-luxembert} and mBERT \cite{devlin2019bertpretrainingdeepbidirectional} on 8 different classification tasks in Luxembourgish. To compare performance with and without orthographic variation, we destandardise or normalise training and test data. Additionally, we fine-tune the models on a combined version of the dataset which includes the standard and non-standard version. 

The results not only show large discrepancies in the performance of models tested on non-standard data but also of models that are trained on data with large amounts of variation. With an extensive understanding of the sociolinguistic situation, we are able to provide a possible solution to include variation in the process and improve the performance with the combined method.

\begin{figure}[ht!]
\centering
\includegraphics[scale=0.18]{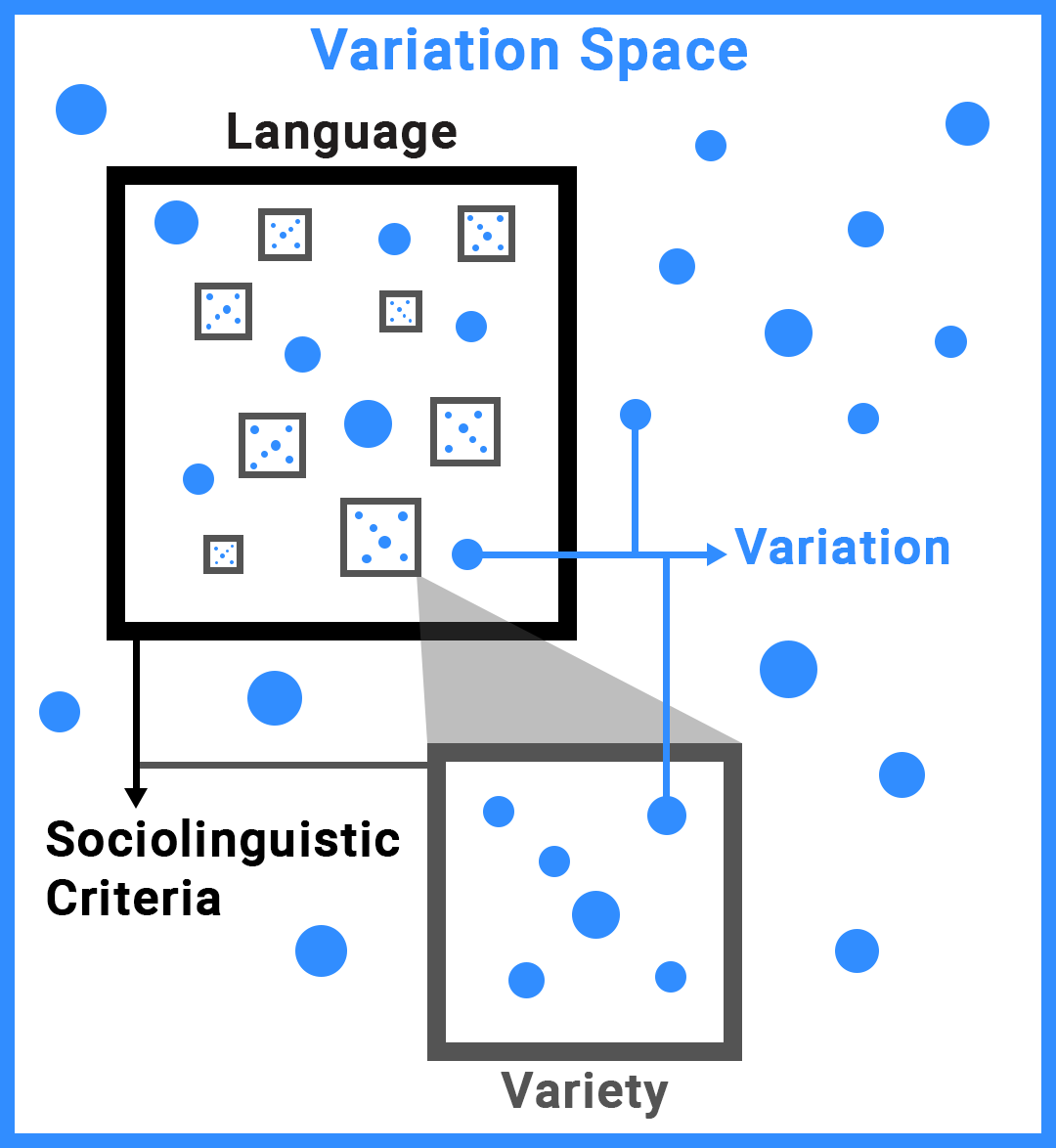}
\caption{Illustration of the container metaphor for language and variety. }
\label{fig:cont}
\end{figure}

\section{Language, Variety, and Variation} \label{definition}
In most contexts, both everyday practice and research, we take the existence of languages as something natural. We differentiate one language from another, and within a language, maybe one dialect from the next. We do so based on our personal experience and the knowledge acquired in practice. From a linguistic point of view, however, what is a language (and what is not) is not evident, but we need to define why we call a particular way of speaking and writing a ``language'' or a ``dialect'' \cite{Cutler_Royneland_Vrzic_2025}. Hence, before we elaborate on the social and technical dimensions of our framework, we need to establish notions of the terms variation, variety and language. In doing so, we mainly draw on variationist literature dealing with regional variation.

\subsection{Defining Varieties and Languages} \label{container}
To understand how we define linguistic entities, we present a simple model of the ``variation space'' from a social interactionist point of view \citep{Sprachdynamik_Auerbook}. Figure~\ref{fig:cont} illustrates the variation space, including linguistic variants (blue points) as well as delimitated containers within this space representing \textit{languages} and \textit{varieties} \citep{Grieve_2025}. To establish these containers, we discuss sociolinguistic criteria based on which varieties and languages can be defined \citep{Purschke_Sprechen}.

The ways in which people speak and write varies in multiple dimensions: regionally, socially, historically, between generations, stylistically, etc. \citep{Gaetano_Auer_book}. The sum of all different ways of speaking and writing defines what we call the ``variation space", i.e., the totality of all existing linguistic variants. Within this space, the continued communication of speakers creates (social) orders of interaction and, hence, orders of linguistic variation. To capture these linguistic orders, \citet{Sprachdynamik_Auerbook} defines three levels of \textit{synchronisation} to reflect the fact that speakers structure the variation space by synchronising their ways of speaking/writing with other speakers (based on their ways of living). On the micro level, individual interactions between speakers create shared (= synchronised) linguistic repertoires, e.g., a couple developing a form of private register shared only between them. Repeated interactions between larger groups of people (social, regional, national, etc.)\ synchronising their ways of speaking/writing define the meso level, i.e., speakers inside the group talk more to each other than they do to outsiders and, hence, develop a shared way of speaking/writing (e.g., a regional dialect). If all speakers within a large community (such as a state) synchronise their communications toward one normative way of speaking/writing, this defines the macro level, best compared to a national or linguistic \textit{norm} (e.g., a standard language). 

Using the concept of synchronisation, we can now understand how speakers (and linguists) delimitate entities within the variation space \cite{Lameli2013}. Starting from the meso level, we assume that continued interactions within groups lead to stabilised ways of speaking/writing distinct from those of other groups, e.g., two neighbouring regional dialects. The distinction between these socially synchronised (internal order) yet (socio)linguistically distinct (external border) ways of speaking/writing invokes the concept of \textit{variety} as used in linguistics \cite{Berruto+2004+293+322}, that is, a way of speaking/writing characteristic for a group of speakers and different from other groups' ways of speaking/writing, based on social, regional, stylistic, etc. criteria. As such, dialects are examples of varieties, as are sociolects or languages.

The question of who defines and labels which linguistic entity in what way (a dialect, a language) depends on the criteria used to delimitate that entity from other entities within the variation space \cite{RambergRoyneland2025}. \citet{Purschke_Sprechen} proposes a list of linguistic systemic and individual subjective criteria by which a group of variants used by a group of speakers (= a socially synchronised way of speaking/writing) can be described and defined \textit{as} a linguistic entity. Those criteria include linguistic differences between two entities, group-related norms and social functions of entities for the linguistic side, and perceived differences, language preferences/attitudes and individual norm concepts for the subjective side. For our framework, we use an adapted version of these (see Section \ref{framework}) as a catalogue of criteria which can be used to describe a linguistic entity based on its (socio)linguistic qualities. In this way, we provide a checklist for NLP purposes to describe and evaluate linguistic entities as cultural constructs, not naturally occurring things.

\subsection{Why is Variation Important?} \label{Var}
By focusing on the standard varieties of languages, in NLP, variation often seems to be a disruptive, `non-standard' factor \cite{plank2016nonstandard}. However, as established in Section \ref{container}, variation is the very fundament of language and social interaction, and the acceptability or correctness of variants are the product of linguistic and social processes of standardisation. Moreover, non-standard variants carry social meaning, i.e., they are indicative of regional, social, national, etc.\ identities \cite{cercas-curry-etal-2024-impoverished}. 

Variation is part of a social semiotic system that conveys the entire range of social concerns of a community \cite{eckert2012three}. Essentially, people express social meaning through variation. Moreover, variation can be a feature indexing ideology, stance or belonging. Even orthographic variation carries not only linguistic meaning (in the sense of correct/incorrect) but also social meaning (in the sense of indexing identities; \citealp{Sebba_2007}). So when people vary their language use, they express who they are, where they belong, and how they want to be seen by others. 

In NLP, variation is often treated as a problem to solve, and as noise in the data to be normalised \cite{eisenstein-2013-bad, al-sharou-etal-2021-towards}. However, normalising language holds the assumption that there exists a default norm which is often not the case, e.g., if we consider non-standardised languages or varieties (such as dialects). Further, normalisation often takes the standard variety and formal register as a default \cite{dogruoz-sitaram-2022-language}, therefore erasing the social meaning from the original text.

\section{Sociolinguistic NLP Framework} \label{framework}
This section introduces our analytical framework. We first discuss nine sociolinguistic criteria of varieties using the example of Luxembourgish in Section \ref{socio}. Then in Section \ref{nlp}, we discuss five essential steps in language modelling and how they relate to the sociolinguistic dimension.

\subsection{Sociolinguistic Criteria} \label{socio}
As discussed in Section \ref{container}, we define varieties/languages as containers within the variation space (Figure \ref{fig:cont}). We now provide a list of sociolinguistic criteria to describe linguistic entities and, hence, delimitate them as varieties/languages based on those criteria. 

\paragraph{The sociolinguistic setting} describes the socio-pragmatic context where this entity is set. This includes multilingualism (individual, societal) and forms of language contact with other entities. 

Luxembourg has three official languages, Luxembourgish (national language), French (legislative language) and German, and is therefore a multilingual society. Luxembourgish has a long history of language contact with German and French as well \cite{gilles_lux_2023}. Additionally, half of the population are foreign residents mostly of French, Portuguese, Italian, Belgian and German descent \cite{STATEC2024}.

\paragraph{Institutional support} describes the political status of the entity and efforts to its societal anchoring via language policy. 

Luxembourgish was established in law in 1984 and is today the national language of the country, following a constitutional reform in 2023. Linguistically, it derives from a Moselle-Franconian dialect. The Centre for the Luxembourgish Language (ZLS), a state run institution, is in charge of the official dictionary and orthography.

\paragraph{Structural independence} describes the structural linguistic differences between the entity and neighbouring ones (e.g., adjacent dialect or roofing standard variety). The degree of linguistic independence from other entities and its linguistic relations to those often define the difference between labelling an entity as a language or a variety of a language \cite{Auer2013}. 
    
Luxembourgish is considered an ``Ausbausprache'' \cite{kloss1967abstand}. In linguistics, it was long seen as a variety of German (Moselle-Franconian), which has developed into a language, covering all functional domains of a standard variety. While there is still regional variation in Luxembourgish, \citet{Gilles1999} finds an advanced state of dialect levelling that resolves the former regional dialects into a national variety with small (lexical, phonological, grammatical) remnants of variation.  

\paragraph{The degree of codification} describes how advanced the orthographic, grammatical and lexical standardisation of an entity is. Standardised languages often represent varieties with high social prestige, act as the main literary language and often are institutionally regulated \cite{bird-2022-local}.

Luxembourgish used to be a mainly spoken language but has gained ground in the written domain in the past 25 years. The language is not fully standardised yet \cite{gilles_lux_2023}, but an official orthography has long existed, last updated in 2019 \cite{ZLS19}. However, since Luxembourgish is not systematically taught in schools, the population has little knowledge of the official rules, resulting in a broad spectrum of variation in written texts \cite{gilles_lux_2023}.

\paragraph{Domain specificity} describes where the entity is used and what functions are attributed to it. 

Luxembourgish multilingualism is characterised by a functional differentiation by social domains. French is the legislative language and important in private business contexts, while public institutions rely on Luxembourgish (parliament, administration). German has its main domains in the traditional print media and in literacy training \cite{purschke2020attitudes}. Luxembourgish is the preferred language in communications among locals and a language of social integration, but given the highly diverse population, the situational choice of a language depends mostly on individual preferences and linguistic repertoires. Luxembourgish is used mostly in informal domains verbally and in writing.

\paragraph{School education} describes the anchoring of an entity in education contexts, that is school curricula or foreign language learning. 

In Luxembourg, German, and lately French, are the official languages for literacy training, with Luxembourgish as an additional language of instruction. In secondary education, German and French serve as the main languages of instruction, although Luxembourgish is informally also present. Nevertheless, there is little to no formal Luxembourgish education at school. There is also high demand for Luxembourgish language courses by immigrants and cross-border workers \cite{Sattler_school}.

\paragraph{Communicative range} describes the size of the speaker group and how useful this language is in social practice. 

Luxembourgish is a small language with around 400.000 speakers and is mostly spoken in Luxembourg, with small pockets in the neighbouring regions due to socio-economic mobility (cross-border commuters) and historical connection (the Belgian region called ``Luxembourg'' that borders Luxembourg). Luxembourgish has a central position as language of social integration but is currently losing ground in the language regime – due to the influx of foreign residents – despite having growing numbers of speakers \cite{Fehlen2023}. In contact with speakers of closely related varieties, i.e., Moselle Franconian in Germany, Luxembourgers most often switch to German, highlighting the cultural difference and, hence, limiting the communicative range of Luxembourgish.

\paragraph{Attitudes and ideologies} are fundamental to social practice, shaping how people use, perceive, and evaluate language \cite{purschke2020attitudes}. Especially in multilingual contexts, attitudes towards different languages reveal much about social dynamics and ideological tensions in a speech community. 
    
In Luxembourg, attitudes towards the different languages reflect the complexity of the societal multilingualism. \Citet{purschke2020attitudes} shows a clear hierarchy of language preferences in daily life, with Luxembourgish being the preferred language in practice for Luxembourgish native speakers. Additionally, there is a close connection between language, 
nation and national identity \cite{Purschke2025}. French is still considered the prestige variety in most contexts, while English is becoming more important. German, while losing ground in the language regime, is preferred over French in  younger generations, also considering its typological closeness with Luxembourgish.

\subsection{NLP Domains} \label{nlp}
If we consider variation to be a part of any variety and therefore a part of language modelling, embracing the sociolinguistic dimension adds a new perspective on known problems in NLP. We have isolated \emph{five distinct domains} in modelling language where variation and different varieties can introduce complexities. In each section we list related work on Luxembourgish as our case study, giving an overview on recent advances.

  \paragraph{Data -- Knowing your data} and its sociolinguistic dimensions allows for a better understanding of the varieties and variation present. For example, \Citet{kreutzer-etal-2022-quality} show that five commonly-used web-crawl corpora (CCAligned, ParaCrawl, WikiMatrix, OSCAR, mC4) are of questionable quality, by identifying languages with no usable text, languages with a very low amount of usable text, and languages with wrong or ambiguous languages codes. Especially low-resource languages are concerned in this quality issue as language identification is especially challenging for language varieties \cite{goot-2025-identifying}. \citet{lau-etal-2025-data} also highlight how common mislabelling in language data is and how this directly influences the performance of language models. 

  For Luxembourgish, automatic language identification is difficult since the language is structurally close to German and exhibits a high amount of French borrowing \cite{gilles_lux_2023}, resulting in low accuracy for language identification. For instance, at first glance, the sentences tagged as Luxembourgish in mC4 \cite{Xue2020mT5AM} and OSCAR \cite{OrtizSuarezSagotRomary2019} are often German, Dutch or French. 
  
  \paragraph{Data -- Data selection} should be influenced by which variety should be represented during training, and not only by the amount of data available. The specific sociolinguistic situation is also an important factor to account for. For instance, \citet{ramponi-2024-language} shows the shortcomings for Italian varieties since NLP focuses more on the amount of resources instead of taking the cultural context into account, like culture preservation, language learning, and intergenerational transmission. He argues for a more responsible speaker-centric approach for varieties in order to preserve language varieties of Italy. The amount of data available is one of many factors to account for while selecting the data for language modelling.
  
  For Luxembourgish, different types of training data exist which are either closer to the orthographic standard or contain a high amount of variation \cite{plum2024textgenerationmodelsluxembourgish}. While selecting the training data, we need to be aware of the content of the training data. A high amount of variation in the training data has an impact on the preprocessing and training, which we show below in Section \ref{exp}.

  \paragraph{Preprocessing -- Unicode character normalisation} specifically for diacritics should be consistent. \citet{gorman-pinter-2025-dont} show that Unicode inconsistencies lead to performance deterioration.

  Diacritics are also widely used in Luxembourgish. They are part of the orthography, specifically to distinguish between different vowel qualities \cite{ZLS19}.  
  
  \paragraph{Preprocessing -- Normalisation} aims to transform non-standard spelling into a standardised form \cite{han-baldwin-2011-lexical,van_der_goot_monoise_2019}. However, by normalising text, we remove rich social signals which are present due to sociolinguistic variation \cite{nguyen-etal-2021-learning}. 

  For Luxembourgish, two normalisation tools are available: \textit{spellux} \cite{purschke2020attitudes},  a normalisation pipeline,  and a neural normalisation model trained with variation-infused parallel data \cite{lutgen-etal-2025-neural}. Both perform similarly, however, \citet{lutgen-etal-2025-neural} argue that normalisation is not effective to smooth training data. In the context of our case study in Section \ref{exp} we show the impact normalisation has on fine-tuning.

  \paragraph{Modelling -- Tokenisation} is sensitive to language variation, and its impact on down-stream tasks depends on the robustness or sensitivity needed to account for variation \cite{wegmann-etal-2025-tokenization}. Subword segmentation for German varieties for instance is challenging as well and does not correspond to a meaningful representation of the data. This is a stark contrast to the German standard variety, as tokenisation difficulties for German varieties lead to low performance in down-stream tasks \cite{blaschke-etal-2023-manipulating}. 

  For Luxembourgish, the impact of data with a high or low amount of variation on tokenisation still needs to be researched. However, we expect differences in tokenisation between the orthographic standard training data and the training data with a high amount of variation. 

  \paragraph{Modelling -- Pre-training strategies} for different varieties must be carefully considered depending on the available data. Modern LLMs are trained on diverse data, including multiple languages and high-noise data, i.e. variation \cite{Grieve_2025}. However, the controlled study of variation for specific, low-resource varieties remains crucial for robustness, fairness, and understanding model limitations. \citet{hedderich-etal-2021-survey} illustrate different strategies for pre-training while working on low-resource languages. One strategy for working with multiple language varieties is adding language labels during training. This enables more cross-lingual transfer as \citet{lample2019crosslinguallanguagemodelpretraining} show for XLM-R and \citet{liu-etal-2020-multilingual-denoising} for mBART. Another strategy is hyper-parameter optimisation based on the amount of training data. The amount of data per language is often regulated (up/downscaled) using a smoothing strategy. Usually, multinomial smoothing is used, where for example XLM-R has a lower upscaling rate \cite{conneau-etal-2020-unsupervised} in comparison to mBERT \cite{devlin2019bertpretrainingdeepbidirectional}. Knowing your training data, the importance of accurate language labels,  the amount of variation in the training data and smoothing parameters impact pre-training strategies. 

  For Luxembourgish, \citet{plum2024textgenerationmodelsluxembourgish} train a T5 model while combining the Luxembourgish training data with French and German data from similar domains. The results show a clear performance boost in comparison to multilingual models, which often use incorrectly tagged non-Luxembourgish training data in the pre-training corpus. Similarly, \textsc{LuxGPT} \cite{bernardy2022} was trained using transfer learning from German. Understanding the prevalence of language contact and the impact on the language with syntactical, lexical and morphological variation make this an informed strategy for upscaling the training data yielding good results. In comparison, \citet{lothritz-etal-2022-luxembert} use synthetic Luxembourgish for data augmentation to train LuxemBERT also yielding good results. 

  \paragraph{Modelling -- Strategies for fine-tuning} for data with a high amount of variation depend on the specific situation and if the data is intended for a classification or generation task. Generally, fine-tuning of pre-trained language models has been shown to be unstable especially for small datasets \cite{du-nguyen-2023-measuring} which is often the case for lower-resource languages. For generative tasks, like neural machine translation, different varieties and variation also pose a challenge. \citet{Zampieri_Nakov_Scherrer_2020} illustrate different strategies used for machine translation for varieties. One strategy, for instance, allows for a shared subword-level vocabulary which enables orthographic and morphological variation learning between related languages. 

  For Luxembourgish, \citet{plum2024textgenerationmodelsluxembourgish} present the generative evaluation dataset LuxGen, which encompasses tasks for standard orthographic and non-standard Luxembourgish. This mix enables them to evaluate models on multiple Luxembourgish varieties. For classification tasks, \citet{lothritz-etal-2022-luxembert} present multiple datasets for classification tasks, however, a fine-grained evaluation on the impact of variation in those tasks is not part of the evaluation. We will show the impact of variation on those classification tasks in Section \ref{exp}. \citet{ranasinghe-etal-2023-publish} benchmark various language models on the task of comment moderation, which encompass a high degree of variation. 

  \paragraph{Evaluation -- Evaluation of non-standard varieties} poses a major challenge as evaluation metrics are not robust to non-standardised varieties, specifically for generation. \citet{aepli-etal-2023-benchmark} show that translation metrics are not reliable for evaluating Swiss German translations, and propose changes to improve robustness. \citet{sun-etal-2023-dialect} illustrate similar findings and conclude that existing metrics for machine translation prioritise dialect similarity over semantics. In both cases, human and automatic evaluations are compared. 

  In Luxembourgish, \citet{plum2024textgenerationmodelsluxembourgish} also evaluate the LuxGen tasks not only with BLEU, as this metric is not suitable for language with a high degree of variation, but also qualitatively. \citet{lutgen-etal-2025-neural} use quantitative and qualitative evaluation metrics to evaluate and compare a ByT5-based normaliser and the normalisation pipeline \textit{spellux} \cite{purschke2020attitudes} for Luxembourgish. By using behavioural performance tests for different correction types, \citet{lutgen-etal-2025-neural} present a fine-grained evaluation on normalisers and show the strengths and weaknesses for each approach.

  \paragraph{Usage -- The usage of language technologies} is becoming more common, therefore the user preferences, especially for lower-resource varieties, should be a concern \cite{markl-etal-2024-language}. For instance, \citet{blaschke-etal-2024-dialect} surveyed German dialect speakers to research language preferences for AI and found that respondents were more interested in potential NLP tools that work with dialectal in- rather than output.

  For Luxembourgish, a similar study has not been performed yet. A study could focus on preferred language use, considering the complex multilingual situation and the varying degrees of adherence to the orthography in the written domain. 
  
  \paragraph{Usage -- Safety issues} for low-resource varieties can be seen in jailbreaking \cite{upadhayay-behzadan-2025-tongue}, as using low-resource languages and languages with a high amount of code switching are effective for jailbreaking attacks. \citet{lent2025weaponizationnlpsecuritymedium} also shows a high security risk for mono- and multilingual models with low-resource languages.

  Safety issues for Luxembourgish have not been researched yet. One could investigate the impact of a high or low degree of variation in the prompts. 

\subsection{How to Use this Framework?}
The goal of the framework we have presented is to systematically include sociolinguistics in NLP. Given the social and linguistic complexity of language, understanding the sociolinguistic perspective on language and how this interacts with language modelling in NLP is essential, both for model performance and a comprehensive representation of linguistic diversity. We argue that sociolinguistic knowledge should inform NLP and, hence, should become an integral part of the research setup. We further demonstrate the usefulness of our framework empirically in a study on orthographic variation in Luxembourgish.

\section{Luxembourgish as a Case Study} \label{exp}
In this section we illustrate how we use the sociolinguistic criteria to inform NLP experimentation. As a case study, we analyse how orthographic variation impacts the performance of fine-tuned models for classification tasks for Luxembourgish. The targeted NLP dimensions are data, preprocessing, modelling (fine-tuning) and evaluation. In the process of analysing variation we have first carefully evaluated the sociolinguistic context, as illustrated in Section \ref{socio}, in order to understand how sociolinguistic criteria impact orthographic variation. The study is informed by the framework's sociolinguistic analysis (Section \ref{framework}) and investigates the impact of variation on specific NLP pipelines.

We consider two varieties of Luxembourgish, the more formal written orthographic standard Luxembourgish called \textbf{standard} and the informal written Luxembourgish that allows for a wide range of orthographic variation called \textbf{non-standard}. Therefore, since we know our data and which variety it represents, we are able to preprocess it for our intended use. We quantitatively evaluate the performance difference of language models for the same downstream tasks in standard Luxembourgish and in non-standard Luxembourgish. Further, we experiment with a \textbf{combined} fine-tuning dataset that encompasses both the standard and the non-standard data of the same datasets to include variation in the training and evaluate the performance. Our fine-tuning setup allows for a fine-grained evaluation of each fine-tuned model (standard, non-standard and combined) on the standard, non-standard  and combined variety of the datasets. By including the sociolinguistic perspective, we not only bring to light the performance differences for different varieties on the same task. We are also able to present a possible solution by combining the standard and non-standard data in order to improve the performance for each evaluated fine-tuned model. 

\subsection{Datasets}\label{datasets}
To evaluate the models on various downstream tasks, we use datasets that are already available for Luxembourgish. Every task is defined as a classification task, either token or sequence based. All the following tasks are from \citetlanguageresource{luxembertdata}, namely: intent classification (IC), winograd natural language inference (WNLI), part-of-speech tagging (POS), named entity classification (NER) and sentiment classification (SC). Additionally topic classification (TC) \citelanguageresource{sib200} and comment moderation (CM) \citelanguageresource{commentdata} are part of our setup. We always use the same hyper-parameters as \citet{lothritz-etal-2022-luxembert} but the data differed for TC as we used \citelanguageresource{sib200} dataset. For CM, we used the same hyperparameter and size of the dataset as for the SC task in \citet{lothritz-etal-2022-luxembert}. More detailed explanation on the tasks including the fine-tuning hyper-parameters can be found in \citetlanguageresource{luxembertdata, sib200, commentdata} and are reported in the Appendix. 

\begin{table}[htbp]
\centering
\scalebox{0.8}{
\begin{tabular}{ll|rr}
\toprule
\textbf{Variant} & \textbf{Task} & \textbf{WER} & \textbf{CER} \\
\midrule
\multirow{5}{*}{\textbf{Destandardised}} 
 & IC & 78.09 & 14.12 \\
 & NER & 63.78 & 12.94 \\
 & POS & 65.61 & 13.00 \\
 & WNLI & 83.23 & 16.02 \\
 & TC & 64.57 & 9.89 \\
\midrule
\multirow{2}{*}{\textbf{Normalised}} 
 & CM & 28.81 & 4.97 \\
 & SC & 28.80 & 4.72 \\
\bottomrule
\end{tabular}
}
\caption{WER (in~\%) and CER (in~\%) between pairs of standard and non-standard variety sentences.}
\label{tab:task_level_metrics}
\end{table}

\subsection{Normalisation \& Destandardisation}
In order to manipulate the datasets of the downstream tasks and inject different amounts of variation into the data, we use normalisation and destandardisation. Normalisation transforms non-standard forms into standard forms with a normalisation pipeline \textit{spellux} \citelanguageresource{purschke2020attitudes2}. Destandardisation instead aims to inject variation, and is performed with the destandardisation algorithm based on \citet{lutgen-etal-2025-neural}. The algorithm is based on data provided by \textit{Spellchecker.lu}\footnote{\url{https://spellchecker.lu}}, a semi-automatic spellchecking website frequently used in Luxembourg. The destandardisation algorithm includes different variants for each word as well as frequency information for user corrections for each variant. This creates a real-life dictionary of spelling variants per lemma, including their frequency of use. This dictionary is then used to replace words with a variant based on the frequency of use. This approach is considered superior to adding random character replacements to the data (generating synthetic data), as this captures real variation patterns in Luxembourgish.

The following tasks are written in the orthographic standard and are thus destandardised to form the non-standard variant of the same task: IC, NER, POS, WNLI and TC. Both SC and CM are based on online comments on news articles and are therefore the non-standard variant by default. SC and CM are normalised to form the standard variant of the task.

Additionally both the standard and non-standard variant of the same datasets were combined to form the \textbf{combined} variant of the same dataset. This data manipulation is performed for the train, test and dev set to fine-tune the models on each variant and evaluate them on each variant. 

To quantify how much variation is introduced, we calculated  word error rate (WER), character error rate (CER) and normalised character of the standard and non-standard variant of the datasets which can be seen in Table \ref{tab:task_level_metrics}. The destandardisation has a mostly uniform pattern and between 60\% and 80\% of the words are changed as the WER describes. However, we can also see that the normalisation process is not changing the data drastically since the WER is around 28\%.

\subsection{Models}
For this experiment, we chose BERT-based models to compare the standard and non-standard version of the classification tasks. Our two models are \textit{LuxemBERT} \cite{lothritz-etal-2022-luxembert} and \textit{mBERT} \cite{devlin2019bertpretrainingdeepbidirectional} as we chose one model trained for Luxembourgish and one multilingual model to compare the performance. Further, encoder-based models are the best choice for classification tasks \cite{weller2025seqvsseqopen, ojo-etal-2025-afrobench}. We repeat the experiments with five random seeds and measure standard deviation and therefore the stability of the fine-tuning process.

\begin{table}[htbp]
\centering
\scriptsize
\setlength{\tabcolsep}{3pt}
\renewcommand{\arraystretch}{1.05}

\begin{minipage}{0.48\textwidth}
\centering
\begin{tabular}{ll|ccc}
\toprule
\textbf{Model} & & std & n-std & comb \\
\midrule
LBERT & std & 57.97 $\pm$ 2.18 & 46.48 $\pm$ 1.75 & 52.60 $\pm$ 0.74 \\
LBERT & n-std & 44.88 $\pm$ 2.27 & 47.67 $\pm$ 2.18 & 46.48 $\pm$ 1.78 \\
LBERT & comb & 68.56 $\pm$ 1.05 & 65.81 $\pm$ 1.64 & 67.22 $\pm$ 1.25 \\
mBERT & std & 26.71 $\pm$ 4.93 & 21.91 $\pm$ 4.49 & 24.64 $\pm$ 4.69 \\
mBERT & n-std & 13.35 $\pm$ 2.82 & 19.07 $\pm$ 2.21 & 16.51 $\pm$ 2.07 \\
mBERT & comb & 45.37 $\pm$ 1.33 & 46.47 $\pm$ 2.72 & 46.02 $\pm$ 2.01 \\
\bottomrule
\end{tabular}
\caption{IC – Weighted F1 (± std)}
\label{tab:IC}
\end{minipage}

\vspace{4pt}

\begin{minipage}{0.48\textwidth}
\centering
\begin{tabular}{ll|ccc}
\toprule
\textbf{Model} & & std & n-std & comb \\
\midrule
LBERT & std & 87.52 $\pm$ 0.04 & 75.73 $\pm$ 0.14 & 81.36 $\pm$ 0.07 \\
LBERT & n-std & 86.90 $\pm$ 0.11 & 83.03 $\pm$ 0.07 & 84.85 $\pm$ 0.08 \\
LBERT & comb & 87.60 $\pm$ 0.08 & 83.03 $\pm$ 0.01 & 85.18 $\pm$ 0.01 \\
mBERT & std & 85.02 $\pm$ 0.52 & 67.68 $\pm$ 0.54 & 76.34 $\pm$ 0.48 \\
mBERT & n-std & 84.61 $\pm$ 0.41 & 81.43 $\pm$ 0.70 & 82.98 $\pm$ 0.55 \\
mBERT & comb & 85.12 $\pm$ 0.06 & 80.84 $\pm$ 0.07 & 82.93 $\pm$ 0.06 \\
\bottomrule
\end{tabular}
\caption{POS – Weighted F1 (± std)}
\label{tab:pos}
\end{minipage}

\vspace{4pt}

\begin{minipage}{0.48\textwidth}
\centering
\begin{tabular}{ll|ccc}
\toprule
\textbf{Model} &  & std & n-std & comb \\
\midrule
LBERT & std & 64.42 $\pm$ 0.57 & 64.08 $\pm$ 0.41 & 64.26 $\pm$ 0.44 \\
LBERT & n-std & 63.14 $\pm$ 0.37 & 64.04 $\pm$ 0.66 & 63.60 $\pm$ 0.41 \\
LBERT & comb & 64.28 $\pm$ 0.05 & 64.54 $\pm$ 0.06 & 64.37 $\pm$ 0.06 \\
mBERT & std & 55.28 $\pm$ 7.39 & 55.41 $\pm$ 7.64 & 55.35 $\pm$ 7.52 \\
mBERT & n-std & 59.38 $\pm$ 1.14 & 60.50 $\pm$ 1.14 & 59.96 $\pm$ 1.06 \\
mBERT & comb & 60.36 $\pm$ 0.09 & 60.68 $\pm$ 0.05 & 60.52 $\pm$ 0.06 \\
\bottomrule
\end{tabular}
\caption{CM – Weighted F1 (± std)}
\label{tab:cm}
\end{minipage}

\label{tab:pos_ic_cm_halfpage}
\end{table}

\subsection{Results}
In the following we discuss the results of three tasks more closely, one sequence classification task (IC), one token classification task (POS) and one sequence classification task for which the normalisation process was used (CM). The results of the remaining tasks are in the Appendix. The results are shown in Table \ref{tab:IC}, \ref{tab:pos} and \ref{tab:cm}. The tables show the weighted F1 score for each model (LuxemBERT and mBERT) trained on the different datasets (standard, destandard and combined) and then tested on the different datasets (standard, destandard and combined). 

First, in nearly all tasks the model fine-tuned exclusively on the non-standard datasets is the worst performing category. In the IC task for instance, shown in Table \ref{tab:IC}, the LuxemBERT model fine-tuned on standard data has an F1 score of 57.97\% for the standard test set but the same model performs a lot worse on the non-standard test set with a score of 46.48\%. The non-standard model variant for the same task also shows a performance drop on the standard test set (44.88\%). Only a slightly better performance for the non-standard test set (47.67\%) is visible even though this model is fine-tuned on the same variant-infused data as the test set. On nearly all tasks, a similar trend as this one can be observed. 

We can also see this trend on the mBERT results. Additionally, the performance is generally lower for all the tasks compared to the LuxemBERT variant, showing  that pre-training with more in-language data is beneficial for the performance of fine-tuned tasks on a specific language. 

Secondly, the fine-tuned combined model is one of the best performing models in nearly all of the tasks. If we look at the IC task again, Table \ref{tab:IC}, the combined variant of LuxemBERT has by far the best performance in all three test sets (standard, non-standard and combined). This shows that including variation in the standard data for the training setup can not only benefit the performance of the model for the non-standard data but also the standard variant. However, Luxembourgish has a high degree of variation in the language even in orthographically standardised text therefore it makes sense that this experimental setup is also beneficial for the standard variant. Yet, the combined fine-tuned model seems to have the most impact on sequence classification tasks. In token classification tasks like POS, shown in Table \ref{tab:pos}, we can only observe a marginal difference. But generally, when variation is part of the training set the performance is higher for the destandard test set. 

Third, the results for both tasks that were normalised for the CM experiment (Table \ref{tab:cm}) show nearly no difference in performance indicating that the current normalisation process is not that successful to make a significant difference. This also aligns with the WER and CER in Table \ref{tab:task_level_metrics}. Although normalisation is typically seen as the solution to work with text including a large amount of variation \cite{van-der-goot-etal-2021-multilexnorm, van-der-goot-cetinoglu-2021-lexical, plank-etal-2020-dan} we can clearly see in our experimental setup that incorporating linguistic variation in addition to the standard form within the training data can yield even more improvement overall in the down-stream tasks. Furthermore, the social meaning that those variants carry are also included in the model training. This results in a more diverse training data and can even be varied depending on the social context of specific tasks.

\section{Conclusion}
This paper proposes a framework to systematically include sociolinguistics in NLP. We show how these sociolinguistic criteria can be used in the case of Luxembourgish and illustrate performance challenges for different varieties of Luxembourgish. By contrasting the orthographic standard and a variation-infused variant of the same data we show how differently the fine-tuned models perform for each variant. Additionally, we show a possible route for improving the performance and robustness by including variation in the training process by combining standard and non-standard data. 

This framework is not only intended for small languages and varieties like Luxembourgish but is universally applicable. The Luxembourgish case study served as a proof-of-concept and the framework is suitable for every language. Given the social and linguistic complexity of language, the sociolinguistic criteria give an informed overview of the variety researched. Our framework could be practically applied in a sort of ``sociolinguistic'' language card in datasets. This would give an extensive overview of the specific varieties included in the dataset and could identify possible problems in an NLP research setup.

\section{Limitations}
This research involves experiments on Luxembourgish language data, where we normalised and destandardised the data. The destandardisation process does not cover the entirety of the variation space in Luxembourg as it only covers the variants included in the data used by the algorithm. We acknowledge that the linguistic coverage of our datasets may not fully reflect the linguistic diversity of Luxembourgish.

\section{Acknowledgements}
This research was supported by the Luxembourg National Research Fund (Project code: C22/SC/117225699) and the ERC Consolidator Grant DIALECT 101043235.

The experiments reported in this paper were conducted on the MeluXina high-performance computing infrastructure, an allocation granted by the University of Luxembourg on the EuroHPC supercomputer hosted by LuxProvide.

We would also like to thank Rob van der Goot, Peter Gilles, Emilia Milano, Felicia Körner, Lou Pepin, Nils Rehlinger and Mélanie Wagner for their invaluable input and Jacques Spedener for the illustration. 

\section{Bibliographical References}\label{sec:reference}

\bibliographystyle{lrec2026-natbib}
\bibliography{lrec2026-example}

\begin{thebibliography}{4}
\expandafter\ifx\csname natexlab\endcsname\relax\def\natexlab#1{#1}\fi

\bibitem[{Adelani et~al.(2023)Adelani, Liu, Shen, Vassilyev, Alabi, Mao, Gao, and Lee}]{sib200}
David Ifeoluwa Adelani and Hannah Liu and Xiaoyu Shen and Nikita Vassilyev and Jesujoba O. Alabi and Yanke Mao and Haonan Gao and Annie En-Shiun Lee. 2023.
\newblock \href {http://arxiv.org/abs/2309.07445} {\emph{SIB-200: A Simple, Inclusive, and Big Evaluation Dataset for Topic Classification in 200+ Languages and Dialects}}.

\bibitem[{Lothritz et~al.(2022)Lothritz, Lebichot, Allix, Veiber, Bissyande, Klein, Boytsov, Lefebvre, and Goujon}]{luxembertdata}
Lothritz, Cedric and Lebichot, Bertrand and Allix, Kevin and Veiber, Lisa and Bissyande, Tegawende and Klein, Jacques and Boytsov, Andrey and Lefebvre, Cl{\'e}ment and Goujon, Anne. 2022.
\newblock \href {https://aclanthology.org/2022.lrec-1.543/} {\emph{{L}uxem{BERT}: Simple and Practical Data Augmentation in Language Model Pre-Training for {L}uxembourgish}}.
\newblock European Language Resources Association.

\bibitem[{Purschke(2020)}]{purschke2020attitudes2}
Christoph Purschke. 2020.
\newblock {Attitudes Toward Multilingualism in Luxembourg. A Comparative Analysis of Online News Comments and Crowdsourced Questionnaire Data}.
\newblock \emph{Frontiers in AI}, 3:536086.

\bibitem[{Ranasinghe et~al.(2023)Ranasinghe, Plum, Purschke, and Zampieri}]{commentdata}
Ranasinghe, Tharindu and Plum, Alistair and Purschke, Christoph and Zampieri, Marcos. 2023.
\newblock \emph{Publish or Hold? {{Automatic}} Comment Moderation in {{Luxembourgish}} News Articles}.

\end{thebibliography}


\begin{thebibliography}{64}
\expandafter\ifx\csname natexlab\endcsname\relax\def\natexlab#1{#1}\fi

\bibitem[{Aepli et~al.(2023)Aepli, Amrhein, Schottmann, and Sennrich}]{aepli-etal-2023-benchmark}
No{\"e}mi Aepli, Chantal Amrhein, Florian Schottmann, and Rico Sennrich. 2023.
\newblock \href {https://doi.org/10.18653/v1/2023.wmt-1.99} {{A Benchmark for Evaluating Machine Translation Metrics on Dialects without Standard Orthography}}.
\newblock In \emph{Proceedings of the Eighth Conference on Machine Translation}.

\bibitem[{Al~Sharou et~al.(2021)Al~Sharou, Li, and Specia}]{al-sharou-etal-2021-towards}
Khetam Al~Sharou, Zhenhao Li, and Lucia Specia. 2021.
\newblock \href {https://aclanthology.org/2021.ranlp-1.7/} {{Towards a Better Understanding of Noise in Natural Language Processing}}.
\newblock In \emph{Proceedings of RANLP}.

\bibitem[{Auer(2013)}]{Auer2013}
Peter Auer. 2013.
\newblock \href {https://doi.org/10.3726/978-3-0352-6311-4} {\emph{{Dialect Divergence at the State Border}}}, pages 295--309. Peter Lang Verlag, Bruxelles, Belgium.

\bibitem[{Bernardy(2022)}]{bernardy2022}
Laura Bernardy. 2022.
\newblock {A Luxembourgish GPT-2 Approach Based on Transfer Learning}.
\newblock Master's thesis, University of Trier.

\bibitem[{Berruto(2004)}]{Berruto+2004+293+322}
Gaetano Berruto. 2004.
\newblock \href {https://doi.org/doi:10.1515/tlir.2004.21.3-4.293} {{The problem of variation}}.
\newblock \emph{The Linguistic Review}, 21(3-4):293--322.

\bibitem[{Berruto(2010)}]{Gaetano_Auer_book}
Gaetano Berruto. 2010.
\newblock \href {https://doi.org/doi:10.1515/9783110220278.226} {\emph{{13. Identifing dimensions of linguistic variation in a language space}}}, pages 226--241. De Gruyter Mouton, Berlin, New York.

\bibitem[{Bird(2022)}]{bird-2022-local}
Steven Bird. 2022.
\newblock \href {https://doi.org/10.18653/v1/2022.acl-long.539} {{Local Languages, Third Spaces, and other High-Resource Scenarios}}.
\newblock In \emph{Proceedings of ACL}.

\bibitem[{Blaschke et~al.(2024)Blaschke, Purschke, Schuetze, and Plank}]{blaschke-etal-2024-dialect}
Verena Blaschke, Christoph Purschke, Hinrich Schuetze, and Barbara Plank. 2024.
\newblock \href {https://doi.org/10.18653/v1/2024.acl-short.74} {{What Do Dialect Speakers Want? A Survey of Attitudes Towards Language Technology for {G}erman Dialects}}.
\newblock In \emph{Proceedings of ACL}.

\bibitem[{Blaschke et~al.(2023)Blaschke, Sch{\"u}tze, and Plank}]{blaschke-etal-2023-manipulating}
Verena Blaschke, Hinrich Sch{\"u}tze, and Barbara Plank. 2023.
\newblock \href {https://doi.org/10.18653/v1/2023.vardial-1.5} {{Does Manipulating Tokenization Aid Cross-Lingual Transfer? A Study on {POS} Tagging for Non-Standardized Languages}}.
\newblock In \emph{Proceedings of VarDial}.

\bibitem[{Cercas~Curry et~al.(2024)Cercas~Curry, Talat, and Hovy}]{cercas-curry-etal-2024-impoverished}
Amanda Cercas~Curry, Zeerak Talat, and Dirk Hovy. 2024.
\newblock \href {https://aclanthology.org/2024.lrec-main.761/} {{Impoverished Language Technology: The Lack of (Social) Class in {NLP}}}.
\newblock In \emph{Proceedings of LREC-COLING}.

\bibitem[{Conneau et~al.(2020)Conneau, Khandelwal, Goyal, Chaudhary, Wenzek, Guzm{\'a}n, Grave, Ott, Zettlemoyer, and Stoyanov}]{conneau-etal-2020-unsupervised}
Alexis Conneau, Kartikay Khandelwal, Naman Goyal, Vishrav Chaudhary, Guillaume Wenzek, Francisco Guzm{\'a}n, Edouard Grave, Myle Ott, Luke Zettlemoyer, and Veselin Stoyanov. 2020.
\newblock \href {https://doi.org/10.18653/v1/2020.acl-main.747} {{Unsupervised Cross-lingual Representation Learning at Scale}}.
\newblock In \emph{Proceedings of ACL}.

\bibitem[{Conneau and Lample(2019)}]{lample2019crosslinguallanguagemodelpretraining}
Alexis Conneau and Guillaume Lample. 2019.
\newblock \href {https://proceedings.neurips.cc/paper_files/paper/2019/file/c04c19c2c2474dbf5f7ac4372c5b9af1-Paper.pdf} {Cross-lingual language model pretraining}.
\newblock In \emph{Advances in Neural Information Processing Systems}, volume~32. Curran Associates, Inc.

\bibitem[{Cutler et~al.(2025)Cutler, R\o{}yneland, and Vrzi\'{c}}]{Cutler_Royneland_Vrzic_2025}
Celia Cutler, Unn R\o{}yneland, and Sebastijan Vrzi\'{c}. 2025.
\newblock \emph{{Language Activism: The Role of Scholars in Linguistic Reform and Social Change}}.
\newblock Cambridge University Press, Cambridge.

\bibitem[{Devlin et~al.(2019)Devlin, Chang, Lee, and Toutanova}]{devlin2019bertpretrainingdeepbidirectional}
Jacob Devlin, Ming-Wei Chang, Kenton Lee, and Kristina Toutanova. 2019.
\newblock \href {http://arxiv.org/abs/1810.04805} {{BERT: Pre-training of Deep Bidirectional Transformers for Language Understanding}}.

\bibitem[{Do{\u{g}}ru{\"o}z and Sitaram(2022)}]{dogruoz-sitaram-2022-language}
A.~Seza Do{\u{g}}ru{\"o}z and Sunayana Sitaram. 2022.
\newblock \href {https://aclanthology.org/2022.sigul-1.12/} {{Language Technologies for Low Resource Languages: Sociolinguistic and Multilingual Insights}}.
\newblock In \emph{Proceedings of SIGUL}.

\bibitem[{Du and Nguyen(2023)}]{du-nguyen-2023-measuring}
Yupei Du and Dong Nguyen. 2023.
\newblock \href {https://doi.org/10.18653/v1/2023.acl-long.342} {{Measuring the Instability of Fine-Tuning}}.
\newblock In \emph{Proceedings of ACL}.

\bibitem[{Eckert(2012)}]{eckert2012three}
Penelope Eckert. 2012.
\newblock \href {https://doi.org/10.1146/annurev-anthro-092611-145828} {{Three Waves of Variation Study: The Emergence of Meaning in the Study of Sociolinguistic Variation}}.
\newblock \emph{Annual Review of Anthropology}, 41:87--100.

\bibitem[{Eckert(2016)}]{Eckert_2016}
Penelope Eckert. 2016.
\newblock \emph{{Variation, meaning and social change}}, page 68–85. Cambridge University Press.

\bibitem[{Eisenstein(2013)}]{eisenstein-2013-bad}
Jacob Eisenstein. 2013.
\newblock \href {https://aclanthology.org/N13-1037/} {{What to do about bad language on the internet}}.
\newblock In \emph{Proceedings of NAACL-HLT}.

\bibitem[{Fehlen et~al.(2023)Fehlen, Gilles, Chauvel, Pigeron-Piroth, Ferro, and Le~Bihan}]{Fehlen2023}
Fernand Fehlen, Peter Gilles, Louis Chauvel, Isabelle Pigeron-Piroth, Yann Ferro, and Etienne Le~Bihan. 2023.
\newblock {RP2021 N\textdegree{}8 -- Linguistic Diversity on the Rise}.
\newblock Technical report, STATEC and University of Luxembourg, Luxembourg.

\bibitem[{Gilles(1999)}]{Gilles1999}
Peter Gilles. 1999.
\newblock \href {https://hdl.handle.net/10993/4339} {\emph{{Dialektausgleich im L\"{e}tzebuergeschen: Zur phonetisch-phonologischen Fokussierung einer Nationalsprache}}}.
\newblock Niemeyer, T\"{u}bingen, Germany.

\bibitem[{Gilles(2023)}]{gilles_lux_2023}
Peter Gilles. 2023.
\newblock \href {https://doi.org/10.1093/acrefore/9780199384655.013.943} {{Luxembourgish}}.
\newblock In Sebastian K\"{u}rschner and Antje Dammel, editors, \emph{Oxford {Encyclopedia} of {Germanic} {Linguistics}}. Oxford University Press, Oxford.

\bibitem[{Gorman and Pinter(2025)}]{gorman-pinter-2025-dont}
Kyle Gorman and Yuval Pinter. 2025.
\newblock \href {https://doi.org/10.18653/v1/2025.naacl-short.25} {{Don{'}t Touch My Diacritics}}.
\newblock In \emph{Proceedings of NAACL-HLT}.

\bibitem[{Grieve et~al.(2025)Grieve, Bartl, Fuoli, Grafmiller, Huang, Jawerbaum, Murakami, Perlman, Roemling, and Winter}]{Grieve_2025}
Jack Grieve, Sara Bartl, Matteo Fuoli, Jason Grafmiller, Weihang Huang, Alejandro Jawerbaum, Akira Murakami, Marcus Perlman, Dana Roemling, and Bodo Winter. 2025.
\newblock \href {https://doi.org/10.3389/frai.2024.1472411} {{The sociolinguistic foundations of language modeling}}.
\newblock \emph{Frontiers in Artificial Intelligence}, Volume 7 - 2024.

\bibitem[{Han and Baldwin(2011)}]{han-baldwin-2011-lexical}
Bo~Han and Timothy Baldwin. 2011.
\newblock \href {https://aclanthology.org/P11-1038} {{Lexical Normalisation of Short Text Messages: Makn Sens a {\#}twitter}}.
\newblock In \emph{Proceedings of ACL}.

\bibitem[{Hedderich et~al.(2021)Hedderich, Lange, Adel, Str{\"o}tgen, and Klakow}]{hedderich-etal-2021-survey}
Michael~A. Hedderich, Lukas Lange, Heike Adel, Jannik Str{\"o}tgen, and Dietrich Klakow. 2021.
\newblock \href {https://doi.org/10.18653/v1/2021.naacl-main.201} {{A Survey on Recent Approaches for Natural Language Processing in Low-Resource Scenarios}}.
\newblock In \emph{Proceedings of NAACL-HLT}.

\bibitem[{Kloss(1967)}]{kloss1967abstand}
Heinz Kloss. 1967.
\newblock {Abstand Languages and Ausbau Languages}.
\newblock \emph{Anthropological Linguistics}, 9(7):29--41.

\bibitem[{Kreutzer et~al.(2022)Kreutzer, Caswell, Wang, Wahab, van Esch, Ulzii-Orshikh, Tapo, Subramani, Sokolov, Sikasote, Setyawan, Sarin, Samb, Sagot, Rivera, Rios, Papadimitriou, Osei, Suarez, Orife, Ogueji, Rubungo, Nguyen, M{\"u}ller, M{\"u}ller, Muhammad, Muhammad, Mnyakeni, Mirzakhalov, Matangira, Leong, Lawson, Kudugunta, Jernite, Jenny, Firat, Dossou, Dlamini, de~Silva, {\c{C}}abuk~Ball{\i}, Biderman, Battisti, Baruwa, Bapna, Baljekar, Azime, Awokoya, Ataman, Ahia, Ahia, Agrawal, and Adeyemi}]{kreutzer-etal-2022-quality}
Julia Kreutzer, Isaac Caswell, Lisa Wang, Ahsan Wahab, Daan van Esch, Nasanbayar Ulzii-Orshikh, Allahsera Tapo, Nishant Subramani, Artem Sokolov, Claytone Sikasote, Monang Setyawan, Supheakmungkol Sarin, Sokhar Samb, Beno{\^i}t Sagot, Clara Rivera, Annette Rios, Isabel Papadimitriou, Salomey Osei, Pedro~Ortiz Suarez, Iroro Orife, Kelechi Ogueji, Andre~Niyongabo Rubungo, Toan~Q. Nguyen, Mathias M{\"u}ller, Andr{\'e} M{\"u}ller, Shamsuddeen~Hassan Muhammad, Nanda Muhammad, Ayanda Mnyakeni, Jamshidbek Mirzakhalov, Tapiwanashe Matangira, Colin Leong, Nze Lawson, Sneha Kudugunta, Yacine Jernite, Mathias Jenny, Orhan Firat, Bonaventure F.~P. Dossou, Sakhile Dlamini, Nisansa de~Silva, Sakine {\c{C}}abuk~Ball{\i}, Stella Biderman, Alessia Battisti, Ahmed Baruwa, Ankur Bapna, Pallavi Baljekar, Israel~Abebe Azime, Ayodele Awokoya, Duygu Ataman, Orevaoghene Ahia, Oghenefego Ahia, Sweta Agrawal, and Mofetoluwa Adeyemi. 2022.
\newblock \href {https://doi.org/10.1162/tacl_a_00447} {{Quality at a Glance: An Audit of Web-Crawled Multilingual Datasets}}.
\newblock \emph{Transactions of ACL}, 10:50--72.

\bibitem[{Lameli(2013)}]{Lameli2013}
Alfred Lameli. 2013.
\newblock \emph{{Strukturen im Sprachraum: Analysen zur arealtypologischen Komplexit\"{a}t der Dialekte in Deutschland}}, volume~54 of \emph{Linguistik – Impulse und Tendenzen}.
\newblock De Gruyter, Berlin, Boston.

\bibitem[{Lau et~al.(2025)Lau, Chen, Fang, Xu, Chen, and Golik}]{lau-etal-2025-data}
Mingfei Lau, Qian Chen, Yeming Fang, Tingting Xu, Tongzhou Chen, and Pavel Golik. 2025.
\newblock \href {https://doi.org/10.18653/v1/2025.acl-long.370} {{Data Quality Issues in Multilingual Speech Datasets: The Need for Sociolinguistic Awareness and Proactive Language Planning}}.
\newblock In \emph{Proceedings of ACL}.

\bibitem[{Lent(2025)}]{lent2025weaponizationnlpsecuritymedium}
Heather Lent. 2025.
\newblock \href {http://arxiv.org/abs/2507.03473} {{Beyond Weaponization: NLP Security for Medium and Lower-Resourced Languages in Their Own Right}}.

\bibitem[{Liu et~al.(2020)Liu, Gu, Goyal, Li, Edunov, Ghazvininejad, Lewis, and Zettlemoyer}]{liu-etal-2020-multilingual-denoising}
Yinhan Liu, Jiatao Gu, Naman Goyal, Xian Li, Sergey Edunov, Marjan Ghazvininejad, Mike Lewis, and Luke Zettlemoyer. 2020.
\newblock \href {https://doi.org/10.1162/tacl_a_00343} {{Multilingual Denoising Pre-training for Neural Machine Translation}}.
\newblock \emph{Transactions of ACL}, 8:726--742.

\bibitem[{Lothritz et~al.(2022)Lothritz, Lebichot, Allix, Veiber, Bissyande, Klein, Boytsov, Lefebvre, and Goujon}]{lothritz-etal-2022-luxembert}
Cedric Lothritz, Bertrand Lebichot, Kevin Allix, Lisa Veiber, Tegawende Bissyande, Jacques Klein, Andrey Boytsov, Cl{\'e}ment Lefebvre, and Anne Goujon. 2022.
\newblock \href {https://aclanthology.org/2022.lrec-1.543/} {{{L}uxem{BERT}: Simple and Practical Data Augmentation in Language Model Pre-Training for {L}uxembourgish}}.
\newblock In \emph{Proceedings of LREC}.

\bibitem[{Lutgen et~al.(2025)Lutgen, Plum, Purschke, and Plank}]{lutgen-etal-2025-neural}
Anne-Marie Lutgen, Alistair Plum, Christoph Purschke, and Barbara Plank. 2025.
\newblock \href {https://aclanthology.org/2025.vardial-1.9/} {{Neural Text Normalization for {L}uxembourgish Using Real-Life Variation Data}}.
\newblock In \emph{Proceedings of VarDial}.

\bibitem[{Markl et~al.(2024)Markl, Hall-Lew, and Lai}]{markl-etal-2024-language}
Nina Markl, Lauren Hall-Lew, and Catherine Lai. 2024.
\newblock \href {https://aclanthology.org/2024.lrec-main.881/} {{Language Technologies as If People Mattered: Centering Communities in Language Technology Development}}.
\newblock In \emph{Proceedings of LREC-COLING}.

\bibitem[{Nguyen et~al.(2021)Nguyen, Rosseel, and Grieve}]{nguyen-etal-2021-learning}
Dong Nguyen, Laura Rosseel, and Jack Grieve. 2021.
\newblock \href {https://doi.org/10.18653/v1/2021.naacl-main.50} {{On learning and representing social meaning in {NLP}: a sociolinguistic perspective}}.
\newblock In \emph{Proceedings of NAACL-HLT}.

\bibitem[{Ojo et~al.(2025)Ojo, Ogundepo, Oladipo, Ogueji, Lin, Stenetorp, and Adelani}]{ojo-etal-2025-afrobench}
Jessica Ojo, Odunayo Ogundepo, Akintunde Oladipo, Kelechi Ogueji, Jimmy Lin, Pontus Stenetorp, and David~Ifeoluwa Adelani. 2025.
\newblock \href {https://doi.org/10.18653/v1/2025.findings-acl.976} {{{A}fro{B}ench: How Good are Large Language Models on {A}frican Languages?}}
\newblock In \emph{Findings of ACL}.

\bibitem[{{Ortiz Su{\'a}rez} et~al.(2019){Ortiz Su{\'a}rez}, Sagot, and Romary}]{OrtizSuarezSagotRomary2019}
Pedro~Javier {Ortiz Su{\'a}rez}, Benoit Sagot, and Laurent Romary. 2019.
\newblock \href {https://doi.org/10.14618/ids-pub-9021} {Asynchronous pipelines for processing huge corpora on medium to low resource infrastructures}.
\newblock Proceedings of the Workshop on Challenges in the Management of Large Corpora (CMLC-7) 2019. Cardiff, 22nd July 2019, pages 9 -- 16, Mannheim. Leibniz-Institut f{"u}r Deutsche Sprache.

\bibitem[{Plank(2016)}]{plank2016nonstandard}
Barbara Plank. 2016.
\newblock What to do about non-standard (or non-canonical) language in nlp.
\newblock In \emph{KONVENS}.

\bibitem[{Plank et~al.(2020)Plank, Jensen, and van~der Goot}]{plank-etal-2020-dan}
Barbara Plank, Kristian~N{\o}rgaard Jensen, and Rob van~der Goot. 2020.
\newblock \href {https://doi.org/10.18653/v1/2020.coling-main.583} {{{D}a{N}+: {D}anish Nested Named Entities and Lexical Normalization}}.
\newblock In \emph{Proceedings of COLING}.

\bibitem[{Plum et~al.(2025)Plum, Ranasinghe, and Purschke}]{plum2024textgenerationmodelsluxembourgish}
Alistair Plum, Tharindu Ranasinghe, and Christoph Purschke. 2025.
\newblock \href {https://aclanthology.org/2025.vardial-1.7/} {{Text Generation Models for Luxembourgish with Limited Data: A Balanced Multilingual Strategy}}.
\newblock In \emph{Proceedings of VarDial}.

\bibitem[{Purschke(2019)}]{Purschke_Sprechen}
Christoph Purschke. 2019.
\newblock \href {https://doi.org/doi:10.1515/9783110571110-002} {\emph{{Vom Sprechen zur Sprache. Versuch \"{u}ber die variationslinguistische Praxis des Begrenzens}}}, pages 9--30. De Gruyter, Berlin, Boston.

\bibitem[{Purschke(2020)}]{purschke2020attitudes}
Christoph Purschke. 2020.
\newblock {Attitudes Toward Multilingualism in Luxembourg. A Comparative Analysis of Online News Comments and Crowdsourced Questionnaire Data}.
\newblock \emph{Frontiers in AI}, 3:536086.

\bibitem[{Purschke(2025)}]{Purschke2025}
Christoph Purschke. 2025.
\newblock \href {https://doi.org/10.5282/jlvs/15} {{Discourse Figures in the Luxembourg Language Debate (2015--2020)}}.
\newblock \emph{Zeitschrift f{\"u}r Sprachvariation und Soziolinguistik}, 1(2):37--53.

\bibitem[{Ramberg and R\o{}yneland(2025)}]{RambergRoyneland2025}
Bj\o{}rn~T. Ramberg and Unn R\o{}yneland. 2025.
\newblock {Norm at play}.
\newblock \emph{Sociolinguistica}, 39(2).

\bibitem[{Ramponi(2024)}]{ramponi-2024-language}
Alan Ramponi. 2024.
\newblock \href {https://doi.org/10.1162/tacl_a_00631} {{Language Varieties of {I}taly: Technology Challenges and Opportunities}}.
\newblock \emph{Transactions of ACL}, 12:19--38.

\bibitem[{Ranasinghe et~al.(2023)Ranasinghe, Plum, Purschke, and Zampieri}]{ranasinghe-etal-2023-publish}
Tharindu Ranasinghe, Alistair Plum, Christoph Purschke, and Marcos Zampieri. 2023.
\newblock {Publish or Hold? {{Automatic}} Comment Moderation in {{Luxembourgish}} News Articles}.
\newblock In \emph{Proceedings of RANLP}.

\bibitem[{Sattler(2021)}]{Sattler_school}
Anna-Sabrina Sattler. 2021.
\newblock \href {http://arxiv.org/abs/https://orbilu.uni.lu/10993/46814} {\emph{{Curriculumentwicklung in einer mehrsprachigen Gesellschaft: Das Beispiel Luxemburg}}}.
\newblock Ph.D. thesis, Unilu - University of Luxembourg, Esch-sur-Alzette, Luxembourg.

\bibitem[{Schmidt(2010)}]{Sprachdynamik_Auerbook}
J\"{u}rgen~Erich Schmidt. 2010.
\newblock \href {https://doi.org/doi:10.1515/9783110220278.201} {\emph{{12. Language and space: The linguistic dynamics approach}}}, pages 201--225. De Gruyter Mouton, Berlin, New York.

\bibitem[{Sebba(2007)}]{Sebba_2007}
Mark Sebba. 2007.
\newblock \emph{{Introduction: society and orthography}}, page 1–9. Cambridge University Press.

\bibitem[{{STATEC}(2024)}]{STATEC2024}
{STATEC}. 2024.
\newblock \emph{{Luxembourg in Figures 2024}}.
\newblock STATEC, Luxembourg City.

\bibitem[{Sun et~al.(2023)Sun, Sellam, Clark, Vu, Dozat, Garrette, Siddhant, Eisenstein, and Gehrmann}]{sun-etal-2023-dialect}
Jiao Sun, Thibault Sellam, Elizabeth Clark, Tu~Vu, Timothy Dozat, Dan Garrette, Aditya Siddhant, Jacob Eisenstein, and Sebastian Gehrmann. 2023.
\newblock \href {https://doi.org/10.18653/v1/2023.acl-long.331} {{Dialect-robust Evaluation of Generated Text}}.
\newblock In \emph{Proceedings of ACL}.

\bibitem[{Upadhayay and Behzadan(2025)}]{upadhayay-behzadan-2025-tongue}
Bibek Upadhayay and Vahid Behzadan. 2025.
\newblock \href {https://doi.org/10.18653/v1/2025.calcs-1.5} {{Tongue-Tied: Breaking {LLM}s Safety Through New Language Learning}}.
\newblock In \emph{Proceedings of the 7th Workshop on Computational Approaches to Linguistic Code-Switching}.

\bibitem[{van~der Goot(2019)}]{van_der_goot_monoise_2019}
Rob van~der Goot. 2019.
\newblock \href {https://doi.org/10.18653/v1/P19-3032} {{{MoNoise}: {A} {Multi}-lingual and {Easy}-to-use {Lexical} {Normalization} {Tool}}}.
\newblock In \emph{Proceedings of ACL ({System} {Demonstrations})}.

\bibitem[{van~der Goot(2025)}]{goot-2025-identifying}
Rob van~der Goot. 2025.
\newblock \href {https://doi.org/10.18653/v1/2025.acl-long.891} {{Identifying Open Challenges in Language Identification}}.
\newblock In \emph{Proceedings of ACL}.

\bibitem[{van~der Goot and {\c{C}}etino{\u{g}}lu(2021)}]{van-der-goot-cetinoglu-2021-lexical}
Rob van~der Goot and {\"O}zlem {\c{C}}etino{\u{g}}lu. 2021.
\newblock \href {https://doi.org/10.18653/v1/2021.eacl-main.200} {{Lexical Normalization for Code-switched Data and its Effect on {POS} Tagging}}.
\newblock In \emph{Proceedings of EACL}.

\bibitem[{van~der Goot et~al.(2021)van~der Goot, Ramponi, Zubiaga, Plank, Muller, San Vicente~Roncal, Ljube{\v{s}}i{\'c}, {\c{C}}etino{\u{g}}lu, Mahendra, {\c{C}}olako{\u{g}}lu, Baldwin, Caselli, and Sidorenko}]{van-der-goot-etal-2021-multilexnorm}
Rob van~der Goot, Alan Ramponi, Arkaitz Zubiaga, Barbara Plank, Benjamin Muller, I{\~n}aki San Vicente~Roncal, Nikola Ljube{\v{s}}i{\'c}, {\"O}zlem {\c{C}}etino{\u{g}}lu, Rahmad Mahendra, Talha {\c{C}}olako{\u{g}}lu, Timothy Baldwin, Tommaso Caselli, and Wladimir Sidorenko. 2021.
\newblock \href {https://doi.org/10.18653/v1/2021.wnut-1.55} {{{M}ulti{L}ex{N}orm: A Shared Task on Multilingual Lexical Normalization}}.
\newblock In \emph{Proceedings of W-NUT 2021}.

\bibitem[{Wang et~al.()Wang, Singh, Michael, Hill, Levy, and Bowman}]{wang-etal-2018-glue}
Alex Wang, Amanpreet Singh, Julian Michael, Felix Hill, Omer Levy, and Samuel Bowman.
\newblock \href {https://doi.org/10.18653/v1/W18-5446} {{GLUE}: A multi-task benchmark and analysis platform for natural language understanding}.
\newblock In \emph{Proceedings of the Workshop {B}lackbox{NLP}: Analyzing and Interpreting Neural Networks for {NLP}}, pages 353--355.

\bibitem[{Wegmann et~al.(2025)Wegmann, Nguyen, and Jurgens}]{wegmann-etal-2025-tokenization}
Anna Wegmann, Dong Nguyen, and David Jurgens. 2025.
\newblock \href {https://doi.org/10.18653/v1/2025.findings-acl.572} {{Tokenization is Sensitive to Language Variation}}.
\newblock In \emph{Findings of ACL}.

\bibitem[{Weller et~al.(2025)Weller, Ricci, Marone, Chaffin, Lawrie, and Durme}]{weller2025seqvsseqopen}
Orion Weller, Kathryn Ricci, Marc Marone, Antoine Chaffin, Dawn Lawrie, and Benjamin~Van Durme. 2025.
\newblock \href {http://arxiv.org/abs/2507.11412} {{Seq vs Seq: An Open Suite of Paired Encoders and Decoders}}.

\bibitem[{Wodak et~al.(2011)Wodak, Johnstone, and Kerswill}]{wodak2011sociolinguistics}
Ruth Wodak, Barbara Johnstone, and Paul Kerswill. 2011.
\newblock \href {https://doi.org/10.4135/9781446200957} {\emph{{The SAGE Handbook of Sociolinguistics}}}.
\newblock SAGE Publications Ltd, London.

\bibitem[{Xue et~al.(2020)Xue, Constant, Roberts, Kale, Al-Rfou, Siddhant, Barua, and Raffel}]{Xue2020mT5AM}
Linting Xue, Noah Constant, Adam Roberts, Mihir Kale, Rami Al-Rfou, Aditya Siddhant, Aditya Barua, and Colin Raffel. 2020.
\newblock \href {https://api.semanticscholar.org/CorpusID:225040574} {mt5: A massively multilingual pre-trained text-to-text transformer}.
\newblock In \emph{North American Chapter of the Association for Computational Linguistics}.

\bibitem[{Zampieri et~al.(2020)Zampieri, Nakov, and Scherrer}]{Zampieri_Nakov_Scherrer_2020}
Marcos Zampieri, Preslav Nakov, and Yves Scherrer. 2020.
\newblock \href {https://doi.org/10.1017/S1351324920000492} {{Natural language processing for similar languages, varieties, and dialects: A survey}}.
\newblock \emph{Natural Language Engineering}, 26(6):595–612.

\bibitem[{{Zenter fir d'L{\"e}tzebuerger Sprooch}(2019)}]{ZLS19}
{Zenter fir d'L{\"e}tzebuerger Sprooch}, editor. 2019.
\newblock \emph{{D'L{\"e}tzebuerger Orthografie}}.
\newblock Zenter fir d'L{\"e}tzebuerger Sprooch, Stroossen.

\end{thebibliography}

\section{Language Resource References}
\label{lr:ref}
\bibliographystylelanguageresource{lrec2026-natbib}
\bibliographylanguageresource{languageresource}

\section{Appendices}
The dataset description for each classification task is presented in Section \ref{data}, with respective sizes and hyperparameter settings, shown respectively in Tables \ref{tab:dataset_sizes} and \ref{tab:para_sizes}. Section \ref{results} contains the results for topic classification, WNLI, NER and sentiment classification (shown in Tables \ref{tab:topic}, \ref{tab:wnli}, \ref{tab:ner} and \ref{tab:sentiment}), which were not presented in the main results section.

\subsection{Datasets} \label{data}

\begin{table}[ht]
\centering
\scalebox{0.8}{%
\begin{tabular}{lccc}
\toprule
Task & Train & Dev & Test \\
\midrule
IC & 698 & 149 & 159 \\
NER & 4298 & 459 & 770 \\
POS & 4278 & 460 & 388 \\
WNLI & 568 & 63 & 136 \\
CM & 6000 & 1000 & 2000 \\
SC & 1299 & 185 & 364 \\
TC & 701 & 99 & 204 \\
\bottomrule
\end{tabular}%
}
\caption{Number of sentences in the training, development, and test sets for each task.}
\label{tab:dataset_sizes}
\end{table}

\begin{table}[ht]
\centering
\scalebox{0.8}{%
\begin{tabular}{lccc}
\toprule
Task & \llap{batch} size & LR & epochs \\
\midrule
IC, WNLI, CM, SC & 16 & 5\textsuperscript{-5} & 5 \\
NER, POS & 16 & 5\textsuperscript{-5} & 3 \\
TC & 16 & 2\textsuperscript{-5} & 5 \\
\bottomrule
\end{tabular}%
}
\caption{Hyperparameter settings for each task, inspired by \citet{lothritz-etal-2022-luxembert}}
\label{tab:para_sizes}
\end{table}

\paragraph{Intent classification (IC)} entails detecting the intent or goal of a text. It is a multi-class classification task that was created by \citet{lothritz-etal-2022-luxembert} and consists of a Banking Client Support Dataset. 
\paragraph{Topic Classification (TC)} is part of the SIB-200 dataset and is annotated with topic labels. The topics include science/technology, travel, politics, sports, health, entertainment and geography \cite{sib200}.
\paragraph{Winograd Natural Language Inference (WNLI)} is an inference task and part of the GLUE benchmark \cite{wang-etal-2018-glue}. \citet{lothritz-etal-2022-luxembert} translated the dataset to Luxembourgish and it consists of two sentences and a label (1 or 0) for each pair. 
\paragraph{Part-of-Speech Tagging (POS)} is a word classification task that assigns a word class to each token. This dataset was annotated automatically with the spaCy pipeline and checked by a human annotator \cite{lothritz-etal-2022-luxembert}. 
\paragraph{Named Entity Recognition (NER)} was also created by \citet{lothritz-etal-2022-luxembert} and is the same dataset as the POS one. It was manually annotated with five labels: Person, Organisation, (natural) Location,
Geopolitical Entity, and Miscellaneous. 
\paragraph{Sentiment Classification (SC)} is a manually annotated subset of online comments on RTL and is annotated by the sentiment of the comment, so either positive, negative or neutral.
\paragraph{Comment Moderation (CM)} is a subset of online comments from RTL, who have manually moderated the comments. The task is to classify whether a comment was archived or published \cite{ranasinghe-etal-2023-publish}.

\subsection{Results} \label{results}
\begin{table}[ht!]
\centering
\scriptsize
\setlength{\tabcolsep}{3pt}
\renewcommand{\arraystretch}{1.05}

\begin{minipage}{0.48\textwidth}
\centering
\begin{tabular}{ll|ccc}
\toprule
\textbf{Model} & & std & n-std & comb \\
\midrule
LBERT & std & 74.75 $\pm$ 2.10 & 66.13 $\pm$ 1.78 & 70.62 $\pm$ 1.39 \\
LBERT & n-std & 61.84 $\pm$ 2.11 & 56.61 $\pm$ 1.54 & 59.59 $\pm$ 1.24 \\
LBERT & comb & 81.16 $\pm$ 1.07 & 77.47 $\pm$ 1.26 & 79.32 $\pm$ 1.14 \\
mBERT & std & 72.74 $\pm$ 6.07 & 66.60 $\pm$ 8.30 & 69.77 $\pm$ 6.91 \\
mBERT & n-std & 75.35 $\pm$ 4.52 & 75.48 $\pm$ 4.47 & 75.43 $\pm$ 4.41 \\
mBERT & comb & 82.74 $\pm$ 0.94 & 82.09 $\pm$ 0.83 & 82.43 $\pm$ 0.57 \\
\bottomrule
\end{tabular}
\caption{TC – Weighted F1 (± std)}
\label{tab:topic}
\end{minipage}

\vspace{4pt}

\begin{minipage}{0.48\textwidth}
\centering
\begin{tabular}{ll|ccc}
\toprule
\textbf{Model} & & std & n-std & comb \\
\midrule
LBERT & std & 50.31 $\pm$ 0.71 & 50.74 $\pm$ 2.22 & 50.57 $\pm$ 1.25 \\
LBERT & n-std & 50.31 $\pm$ 0.71 & 50.74 $\pm$ 2.22 & 50.57 $\pm$ 1.25 \\
LBERT & comb & 50.31 $\pm$ 0.71 & 50.74 $\pm$ 2.22 & 50.57 $\pm$ 1.25 \\
mBERT & std & 52.66 $\pm$ 2.80 & 50.49 $\pm$ 1.15 & 52.10 $\pm$ 1.47 \\
mBERT & n-std & 52.80 $\pm$ 2.48 & 51.38 $\pm$ 1.83 & 52.30 $\pm$ 1.52 \\
mBERT & comb & 54.86 $\pm$ 3.46 & 52.00 $\pm$ 3.70 & 54.06 $\pm$ 2.91 \\
\bottomrule
\end{tabular}
\caption{WNLI – Weighted F1 (± std)}
\label{tab:wnli}
\end{minipage}

\label{tab:topic_wnli_halfpage}

\vspace{4pt}

\begin{minipage}{0.48\textwidth}
\centering
\begin{tabular}{ll|ccc}
\toprule
\textbf{Model} & & std & n-std & comb \\
\midrule
LBERT & std & 65.62 $\pm$ 0.00 & 60.10 $\pm$ 0.01 & 62.84 $\pm$ 0.00 \\
LBERT & n-std & 65.00 $\pm$ 0.00 & 62.41 $\pm$ 0.00 & 63.67 $\pm$ 0.00 \\
LBERT & comb & 65.94 $\pm$ 0.03 & 63.52 $\pm$ 0.03 & 64.71 $\pm$ 0.02 \\
mBERT & std & 68.23 $\pm$ 0.00 & 55.82 $\pm$ 0.00 & 61.92 $\pm$ 0.00 \\
mBERT & n-std & 68.60 $\pm$ 0.01 & 64.84 $\pm$ 0.02 & 66.71 $\pm$ 0.01 \\
mBERT & comb & 66.82 $\pm$ 0.01 & 63.81 $\pm$ 0.01 & 65.35 $\pm$ 0.01 \\
\bottomrule
\end{tabular}
\caption{NER – Weighted F1 (± std)}
\label{tab:ner}
\end{minipage}

\vspace{4pt}

\begin{minipage}{0.48\textwidth}
\centering
\begin{tabular}{ll|ccc}
\toprule
\textbf{Model} & & std & n-std & comb \\
\midrule
LBERT & std & 62.37 $\pm$ 0.66 & 62.40 $\pm$ 0.53 & 62.40 $\pm$ 0.52 \\
LBERT & n-std & 62.94 $\pm$ 0.50 & 64.04 $\pm$ 0.83 & 63.49 $\pm$ 0.48 \\
LBERT & comb & 63.37 $\pm$ 0.04 & 64.13 $\pm$ 0.13 & 63.75 $\pm$ 0.08 \\
mBERT & std & 55.78 $\pm$ 1.53 & 54.51 $\pm$ 1.24 & 55.16 $\pm$ 1.36 \\
mBERT & n-std & 55.15 $\pm$ 1.14 & 55.55 $\pm$ 1.18 & 55.36 $\pm$ 1.07 \\
mBERT & comb & 56.83 $\pm$ 0.15 & 57.73 $\pm$ 0.29 & 57.29 $\pm$ 0.20 \\
\bottomrule
\end{tabular}
\caption{SC – Weighted F1 (± std)}
\label{tab:sentiment}
\end{minipage}

\label{tab:ner_sentiment_halfpage}
\end{table}

\end{document}